\documentclass{article}

     \usepackage[final]{neurips_2019}

\usepackage[utf8]{inputenc} 
\usepackage[T1]{fontenc}    
\usepackage{url}            
\usepackage{booktabs}       
\usepackage{amsfonts}       
\usepackage{nicefrac}       
\usepackage{microtype}      
\usepackage{graphicx}
\usepackage{subcaption}
\usepackage{amsmath}
\usepackage{bm}
\usepackage{bbm}
\usepackage{algorithmic}
\usepackage{algorithm}
\usepackage{tabularx}
\usepackage{amssymb,amsthm}

\usepackage[font=small,labelfont=bf]{caption}

\usepackage[usenames,dvipsnames]{xcolor}
\definecolor{shadecolor}{gray}{0.9}

\usepackage[colorlinks,linktoc=all]{hyperref}
\hypersetup{citecolor=MidnightBlue}
\hypersetup{linkcolor=MidnightBlue}
\hypersetup{urlcolor=MidnightBlue}
\usepackage{url}

\newcommand{\Real}[0]{\mathbb{R}}

\title{Gradient-based Adaptive Markov Chain Monte Carlo}

\author{
  Michalis K. Titsias\\
        DeepMind \\
       London, UK \\
      \texttt{mtitsias@google.com} 
   \And
	Petros Dellaportas\\
	Department of Statistical Science\\
	University College of London, UK\\
	Department of Statistics, Athens \\
	Univ. of Econ. and Business, Greece\\
	and The Alan Turing Institute, UK
}
\begin{document}

\maketitle

\begin{abstract}
 We introduce a gradient-based learning method to automatically adapt Markov chain Monte Carlo (MCMC) proposal distributions to intractable targets. We define a maximum entropy regularised objective function,  referred to as generalised speed measure, 
which can be robustly optimised over the parameters of the proposal distribution by applying stochastic gradient 
optimisation. An advantage of our method compared to traditional adaptive MCMC methods is that the adaptation occurs even when candidate state values are rejected.  This is a highly desirable property of any adaptation strategy because the adaptation starts in early iterations even if the initial proposal distribution is far from optimum. 
We apply the framework for learning multivariate random walk Metropolis and Metropolis-adjusted Langevin proposals with full covariance matrices,  
  and provide empirical evidence 
  that our method can outperform other MCMC algorithms, including Hamiltonian 
 Monte Carlo schemes. 
\end{abstract}

\section{Introduction}
\label{sec:introduction}

Markov chain Monte Carlo (MCMC) is a family of algorithms that provide a mechanism for generating dependent draws from arbitrarily complex distributions.  The basic set up of an MCMC algorithm in any probabilistic (e.g.\ Bayesian) inference problem,  with an intractable target 
density $\pi(x)$, 
 is  as follows.   A discrete time Markov chain $\{X_t\}_{t=0}^\infty$ with transition kernel $P_\theta$,  appropriately chosen from a collection of $\pi$-invariant kernels 
$\{P_\theta(\cdot,\cdot)\}_{\theta \in \Theta}$, is generated and the ergodic averages 
$
\mu_t(F) = t^{-1} \sum_{i=0}^{t-1} F(X_i) 
$ 
are used as approximations to  $E_\pi(F)$  
for any real-valued function $F$. 
 Although in principle this sampling setup is simple, the actual implementation of any MCMC algorithm requires careful choice of $P_\theta$ because the properties of $\mu_t$ depend on $\theta$.  In common kernels that lead to random walk Metropolis  (RWM), Metropolis-adjusted Langevin (MALA) or Hamiltonian Monte Carlo (HMC) algorithms  the kernels $P_\theta$ are specified through an accept-reject mechanism in which the chain moves from time $t$ to time $t+1$ by first proposing candidate values $y$ from a density $q_\theta(y|x)$ and accepting them with some probability $\alpha(x_t,y)$ and setting $x_{t+1}=y$, or rejecting them and setting $x_{t+1}=x_t$.  Since $\theta$ directly affects this acceptance probability, it is  clear that one should choose $\theta$ so that the chain does not move too slowly or rejects too many proposed values $y$ because in both these cases convergence to the stationary distribution will be slow.  This has been recognised as early as in \citep{metropolis1953equation} and has initiated exciting research that has produced optimum average acceptance probabilities 
for a series of algorithms; see \citep{roberts1997weak,  roberts1998optimal, roberts2001optimal, haario2005componentwise, bedard2007weak, bedard2008optimal, roberts2009examples, bedard2008efficient,  rosenthal2011optimal, beskos2013optimal}.  
Such optimal average acceptance probabilities provide basic guidelines for adapting single 
step size parameters to achieve certain average acceptance rates. 

More sophisticated adaptive MCMC algorithms that can learn a full set of parameters $\theta$, such as a  covariance matrix,
borrow information from the history of the chain to optimise 
some criterion reflecting the performance of the Markov chain 
\citep{haario2001adaptive, atchade2005adaptive, roberts2007coupling, giordani2010adaptive, andrieu2006ergodicity, andrieu2007efficiency, atchade2009adaptive}. Such methods are typically non gradient-based and the basic strategy they use 
is to sequentially fit the proposal $q_\theta(y|x)$ to the history of states $x_{t-1}, x_t,\ldots,$ by ignoring also the 
rejected state values. This can result in very slow adaptation because the initial Markov chain simulations  
are based on poor initial $\theta$ and the generated states, from which $\theta$ is learnt, 
are highly correlated and far from the target. The authors in \cite{roberts2009examples} call such adaptive strategies `greedy' in the sense that they try to adapt too closely to initial information from the output and take considerable time to recover from misleading initial information.

In this paper, we develop faster and more robust gradient-based adaptive MCMC algorithms that make use of the gradient of the 
target, $\nabla \log \pi(x)$,  
and they learn from both actual states of the chain and proposed (and possibly rejected) states. 
The key idea is to define and maximise w.r.t.\  $\theta$ an entropy regularised objective function that promotes
high acceptance rates and high values for the entropy of the proposal distribution.  This objective function, referred to as generalised 
speed measure,  is inspired by the speed measure of the infinite-dimensional limiting diffusion process that 
captures the notion of speed in which a Markov chain converges to its stationary distribution \citep{roberts2001optimal}. 
We maximise this objective function by applying stochastic gradient variational inference techniques such as those 
based on the reparametrisation trick \citep{Kingma2014,Rezende2014,Titsias2014_doubly}.
 An advantage of our algorithm compared to traditional adaptive MCMC methods is that the adaptation occurs even when candidate state values are rejected.  In fact, the adaptation can be faster when candidate values $y$ are rejected since then we make always 
 full use of the gradient $\nabla \log \pi(y)$ evaluated at the rejected $y$. This allows 
 the adaptation to start in early iterations even if the initial proposal distribution is far from optimum 
 and the chain is not moving.  We apply the method for learning multivariate RWM and MALA
  proposals where we adapt  full covariance matrices parametrised efficiently using Cholesky factors.    
  In the experiments we demonstrate our algorithms to multivariate Gaussian targets and Bayesian logistic regression and 
  empirically show that they outperform several other baselines, including advanced HMC schemes.

\section{Gradient-based adaptive MCMC}
\label{sec:theory}

Assume a target distribution $\pi(x)$, known up to some unknown normalising constant, where 
$x \in \Real^n $ is the state vector. 
To sample from $\pi(x)$ we consider the Metropolis-Hastings (M-H) algorithm 
that generates new states  
by sampling from a proposal distribution $q_{\theta}(y|x)$, that depends on 
parameters $\theta$, and accepts or rejects each proposed state by using the standard M-H
acceptance probability
\begin{equation}
  \alpha(x,y;\theta) = \text{min}\left\{1,
  \frac{ \pi(y) q_{\theta}(x|y)}
       { \pi(x) q_{\theta}(y|x)}  
  \right\}.
\end{equation}
While the M-H algorithm defines a Markov chain that converges to the
target distribution, the efficiency of the algorithm heavily depends on the choice
of the proposal distribution $q_{\theta}(x|y)$ and the setting of its parameters 
$\theta$.  

Here, we develop a framework for stochastic gradient-based adaptation or learning of 
$q_{\theta}(x|y)$ that maximises  an objective
function inspired by the concept of speed measure that
underlies the theoretical foundations of MCMC  optimal tuning
\citep{roberts1997weak,  roberts1998optimal}. Given that the chain is at state $x$
we would like: (i) to propose big jumps in the state space
and (ii) accept these jumps with high probability.  By assuming for now that the 
proposal has the standard random walk isotropic form, such 
that $q_{\sigma}(y|x) =  \mathcal{N}(y|x, \sigma^2 I )$, 
the speed measure is defined as 
\begin{equation}
s_{\sigma}(x) =  \sigma^2 \times \alpha(x;\sigma),
  \label{eq:speedmeasure}
\end{equation}
where $\sigma^2$ denotes the variance, also called step size,  of the proposal distribution,
while $\alpha(x; \sigma)$ is the average acceptance probability 
when starting at $x$, i.e.\  $ \alpha(x; \sigma) = \int  \alpha(x,y; \sigma) q_{\sigma} (y | x) d y$.  To learn a good value for the 
step size we could maximise the speed measure in Eq.\ \ref{eq:speedmeasure},  
which intuitively promotes high variance for the proposal distribution together with high acceptance rates. 
In the theory of optimal MCMC tuning, $s_{\sigma}(x)$ is averaged under the stationary distribution $\pi(x)$ 
to obtain a global speed measure value $s_{\sigma} = \int \pi(x) s_{\sigma}(x) d x$.
For simple targets and with increasing dimension this measure  
is maximised so that $\sigma^2$ is set to a value that leads to the acceptance probability  
$0.234$ \citep{roberts1997weak,  roberts1998optimal}. This subsequently leads to the popular 
heuristic for tuning random walk proposals: tune $\sigma^2$ so that 
on average the proposed states are accepted with probability $1/4$.  Similar heuristics have been obtained for tuning 
the step sizes of more advanced  schemes such as MALA and HMC, 
where 0.574 is considered optimal for MALA  \citep{roberts2001optimal} 
and 0.651 for HMC \citep{neal2010,beskos2013optimal}.      

While the current notion of speed measure from Eq.\  \ref{eq:speedmeasure} is intuitive, it 
is only suitable for tuning proposals having a single step size. Thus,
in order to learn arbitrary proposal distributions $q_{\theta}(y |x)$, where 
$\theta$ is a vector of parameters, we need to define suitable generalisations of  the speed measure. 
Suppose, for instance,  that we wish to tune 
a Gaussian with a full covariance matrix, i.e.\
$q_{\Sigma}(y|x) = \mathcal{N}(y|x, \Sigma)$. To achieve this
we need to generalise the objective  in Eq.\  \ref{eq:speedmeasure} by  
replacing  $\sigma^2$ with some functional $\mathcal{F}(\Sigma)$  that depends 
on the full covariance.  An obvious choice is to 
consider the average distance  $||y-x||^2$ given 
by the trace $\text{tr}(\Sigma) = \sum_{i} \sigma_i^2$. However, this is problematic  
since it can lead to learning proposals with very poor mixing. 
To see this note that since  the trace is the sum of variances it  
can obtain high values even when some of the components of $x$ have very low variance,  
e.g.\ for some $x_i$  it holds $\sigma_i^2 \approx 0$, which can result in very low 
sampling efficiency or even non-ergodicity.  In order to define better functionals $\mathcal{F}(\Sigma)$ 
we wish to exploit the intuition that for MCMC all components of $x$ need to jointly perform (relative to their scale)
big jumps, a requirement that is better captured by the determinant $|\Sigma|$ or 
more generally by the entropy of the proposal distribution.

Therefore, we introduce a generalisation of the speed measure having the form, 
 \begin{equation}
  s_{\theta}(x) = \exp\{ \beta \mathcal{H}_{q_{\theta}(y | x)} \} \times \alpha(x; \theta) 
  =  \exp\{ \beta \mathcal{H}_{q_{\theta}(y | x)} \} \times \int \alpha(x,y; \theta) q_{\theta}(y | x) d y ,
 \label{eq:speedmeasure2}
 \end{equation}
 where $ \mathcal{H}_{q_{\theta}(y | x)}  = - \int  q_{\theta} (y | x) \log q_{\theta}(y|x) d y$  
 denotes the entropy of the proposal distribution and  $\beta > 0$ is an hyperparameter. Note that when the 
 proposal distribution  is a full Gaussian $q_{\Sigma}(y|x) = \mathcal{N}(y|x, \Sigma)$
 then $\exp\{ \beta \mathcal{H}_{q(y | x)} \} = \text{const} \times |\Sigma|^{ \frac{\beta}{2}}$ which depends
 on the determinant of $\Sigma$.  $s_{\theta}(x)$, referred to as \emph{generalised speed measure}, 
 trades off between high entropy of the proposal distribution and high acceptance probability.     
 The hyperparameter $\beta$ plays the crucial role of  balancing the relative 
 strengths of these terms.  As discussed in the next section we can efficiently optimise 
 $\beta$ in order to achieve a desirable average acceptance rate.     
 
 In the following we make use of the above generalised speed measure 
 to derive a variational learning algorithm for adapting the parameters 
 $\theta$  using stochastic gradient-based optimisation.


 \subsection{Maximising the generalised speed measure using variational inference \label{sec:maximisespeed}}

 During MCMC iterations we collect the pairs of vectors $(x_t,y_t)_{t>0}$ 
 where $x_t$ is the state of the chain at time $t$ and $y_t$ the corresponding proposed next state, which if accepted then $x_{t+1}=y_t$. 
 When the chain has converged each $x_t$ follows the stationary distribution $\pi(x)$, otherwise it follows
  some distribution that progressively converges to $\pi(x)$. In either case we view the sequence of pairs $(x_t,y_t)$ as 
  non-iid data based on which we wish to perform gradient-based updates 
 of the parameters $\theta$. In practice such updates can be performed with diminishing learning 
 rates, or more safely completely stop after some number of burn-in iterations to ensure convergence.  
 Specifically, given the current state $x_t$ 
 we wish to take a step towards maximising $s_{\theta}(x_t)$ in Eq.\ \ref{eq:speedmeasure2} or equivalently 
 its logarithm, 
\begin{equation}
\log s_{\theta}(x_t) = \log \int \alpha(x,y; \theta) q_{\theta}(y | x_t) d y   + \beta \mathcal{H}_{q_{\theta}(y | x_t)}.
\end{equation}
The second term is just the entropy of the proposal distribution, which typically will be analytically tractable, 
while the first term involves an intractable integral.
To approximate the first term we work similarly to variational inference  \citep{Jordan1999learn,Bishop2006} 
and we lower bound it using Jensen's inequality,
\begin{align}
 \log s_{\theta}(x_t) \geq   \mathcal{F}_{\theta} (x_t)  & =   \int q_{\theta}(y | x_t)  \log 
 \text{min}\left\{ 1,
  \frac{ \pi(y) q_{\theta}(x_t|y)}
       { \pi(x_t) q_{\theta}(y|x_t)}   \right\}  d y  + \beta \mathcal{H}_{q_{\theta}(y | x_t)} \\
& = \int q_{\theta} (y | x_t)
  \text{min}\left\{0, \log \frac{ \pi(y) }{\pi(x_t)}  + \log \frac{q_{\theta}(x_t | y)}{q_{\theta}(y | x_t)} \right\} d y  +  \beta \mathcal{H}_{q_{\theta}(y | x_t)}. 
\end{align}
To take a step towards maximising $\mathcal{F}_{\theta}$ we can apply standard stochastic variational 
inference techniques such as the score function method or the reparametrisation trick
\citep{Carbonetto2009,Paisley2012,Ranganath2014,Kingma2014,Rezende2014,Titsias2014_doubly,Kucukelbir2017}.   
Here, we focus on the case  $q_{\theta}(y | x_t)$ is a reparametrisable 
distribution such that $y = \mathcal{T}_{\theta}(x_t, \epsilon)$ where $\mathcal{T}_{\theta}$ is a deterministic transformation
and $\epsilon \sim p(\epsilon)$.  
$\mathcal{F}_{\theta}(x_t)$ can be re-written as
\begin{align}
 \mathcal{F}_{\theta} (x_t) & =  \int p(\epsilon)
  \text{min}\left\{0,
  \log \frac{\pi( \mathcal{T}_{\theta} (x_t, \epsilon) )}{\pi(x_t)}
  + \log \frac{q_{\theta}(x_t | \mathcal{T}_{\theta} (x_t, \epsilon))}{ q_{\theta}( \mathcal{T}_{\theta} (x_t, \epsilon ) | x_t)}
\right\} d \epsilon \nonumber + \beta \mathcal{H}_{q_{\theta}(y | x_t)}.
\end{align} 
Since MCMC at the $t$-th iteration proposes a specific state $y_t$ constructed as $\epsilon_t \sim p(\epsilon_t)$, 
$y_t = \mathcal{T}_{\theta} (x_t, \epsilon_t)$, an unbiased estimate of the exact gradient 
$\nabla_{\theta} \mathcal{F}_{\theta} (x_t)$ can be obtained by    
\begin{align}
\nabla_{\theta} \mathcal{F}_{\theta} (x_t, \epsilon_t) = 
 \nabla_{\theta}   \text{min}\left\{ 0,
  \log \frac{\pi( \mathcal{T}_{\theta} (x_t, \epsilon_t ) )}{\pi(x_t)}
  + \log \frac{q_{\theta}(x_t | \mathcal{T}_{\theta} (x_t, \epsilon_t))}{ q_{\theta}( \mathcal{T}_{\theta} (x_t, \epsilon_t ) | x_t)}
\right\}  \nonumber + \beta \nabla_{\theta} \mathcal{H}_{q_{\theta}(y | x_t)},
\end{align}
which is used to make a gradient update for the parameters $\theta$.
Note that the first term 
 in the stochastic gradient is analogous to differentiating through 
a  rectified linear hidden unit (ReLu) in neural networks, 
  i.e. if  $\log \frac{\pi(y_t)}{\pi(x_t)}  +\log \frac{q_{\theta}(x_t | y_t)}{ q_{\theta}( y_t | x_t)} \geq 0$ the gradient is zero 
 (this is the case when $y_t$ is accepted with probability one), while otherwise the gradient of the first term reduces to 
 $$
\nabla_{\theta}  \log \pi( \mathcal{T}_{\theta} (x_t, \epsilon_t) )
  + \nabla_{\theta} \log \frac{q_{\theta}(x_t | \mathcal{T}_{\theta} (x_t, \epsilon_t))}{ q_{\theta}( \mathcal{T}_{\theta} (x_t, \epsilon_t) | x_t)}.
 $$
The value of the hyperparameter $\beta$ can allow to trade off between large acceptance probability and  
large entropy of the proposal distribution.   
Such hyperparameter cannot be optimised by maximising the  variational objective $\mathcal{F}_{\theta}$ 
(this typically will set  $\beta$  to a very small value so that the acceptance probability becomes close to
 one but the chain is not moving since the entropy is very low). 
 Thus, $\beta$  needs to be updated in order to control the average 
 acceptance probability of the chain in order to achieve a certain desired value $\alpha_*$. 
 The value of $\alpha_*$ can be determined based on the specific MCMC proposal we 
 are using and by following standard recommendations in the literature, as discussed also in the previous section. 
 For instance,  when we use RWM $\alpha_*$ can be set to  
 $1/4$ (see Section \ref{sec:gadRWM}), while for gradient-based MALA  (see Section \ref{sec:gadMALA})
 $\alpha_*$ can be set to $0.55$. 
 
Pseudocode for the general procedure is given by Algorithm 1.    
We set the learning rate $\rho_t$ using RMSProp \citep{Tieleman2012}; 
at each iteration $t$ we set $\rho_t =\eta /(1+\sqrt{G_t } )$, where $\eta$ is the baseline 
learning rate, and the updates of $G_t$ depend on the gradient estimate $\nabla_{\theta} \mathcal{F}_{\theta}(x_t, \epsilon_t)$ 
as $G_t = 0.9G_t + 0.1 \left[ \nabla_{\theta} \mathcal{F}_{\theta}(x_t, \epsilon_t) \right]^2$.

\begin{algorithm}[tb]
   \caption{Gradient-based Adaptive MCMC}
   \label{alg:example}
\begin{algorithmic}
   \STATE {\bfseries Input:} target $\pi(x)$; reparametrisable proposal $q_{\theta}(y|x)$ s.t.\  
   $ y = \mathcal{T}_{\theta} (x, \epsilon)$, $\epsilon \sim p(\epsilon)$; initial $x_0$; desired average acceptance probability $\alpha_{*}$.    
   \STATE Initialise $\theta$, $\beta=1$.
   \FOR{$t=1,2,3,\ldots,$}
   \STATE: Propose $\epsilon_t \sim p(\epsilon_t)$, $y_t =  \mathcal{T}_{\theta} (x_t, \epsilon_t)$.
   \STATE: Adapt $\theta$: $\theta \leftarrow \theta + \rho_t \nabla_{\theta} \mathcal{F}_{\theta} (x_t, \epsilon_t)$. 
   \STATE: Accept or reject $y_t$ using the standard M-H ratio to obtain $x_{t+1}$. 
   \STATE: Set $\alpha_t=1$ if $y_t$ was accepted and $\alpha_t=0$ otherwise.
   \STATE: Adapt hyperparameter $\beta$:  $\beta \leftarrow \beta [1 + \rho_{\beta} (\alpha_t - \alpha_*) ]$  \# default value for $\rho_{\beta} = 0.02$.
   \ENDFOR
\end{algorithmic}
\end{algorithm}

\subsection{Fitting a full covariance Gaussian random walk proposal \label{sec:gadRWM}} 

We now specialise to the case  the proposal distribution is a random walk Gaussian  
$q_{L}(y  | x ) = \mathcal{N}(y | x, L L^\top )$ where the parameter $L$ is a positive definite   
 lower triangular matrix, i.e.\ a Cholesky factor. This distribution is reparametrisable since $y \equiv \mathcal{T}_L (x, \epsilon) = x + L \epsilon$, $\epsilon \sim \mathcal{N}(\epsilon | 0, I)$. 
 At the $t$-th iteration when the state is $x_t$ the lower bound becomes 
 \begin{align}
\mathcal{F}_{L}(x_t) 
 & = \int \mathcal{N}(\epsilon | 0, I)  \text{min}\left\{ 
 0, \log \pi(x_t + L \epsilon) - \log \pi(x_t)
 \right\} d \epsilon  +  \beta \sum_{i=1}^n \log L_{i i} + \text{const}.  
 \end{align}  
Here, the proposal distribution has cancelled out from the M-H ratio, since it is symmetric, while
 $ \mathcal{H}_{q_{\theta}(y | x_t)} = \log | L | + \text{const}$ and $\log |L| = \sum_{i=1}^n \log L_{i i}$. 
By making use of the MCMC proposed state $y_t = x_t + L \epsilon_t$ we can obtain an unbiased estimate of the exact gradient  
$\nabla_L \mathcal{F}_{L}(x_t)$,
 \[
\nabla_{L} \mathcal{F}_{L}(x_t, \epsilon_t) =
\begin{cases}
\left[ \nabla_{y_t} \log \pi(y_t) \times \epsilon_t^\top\right]_{lower}  + \beta \text{diag}( \frac{1}{L_{1 1}}, \ldots, \frac{1}{L_{n n}}), &  \text{if}  \ \log \pi(y_t) < \log \pi(x_t) \\
\beta \text{diag}(\frac{1}{L_{1 1}}, \ldots, \frac{1}{L_{n n}}) , &  \text{otherwise} 
\end{cases}
\]
where $y_t = x_t + L \epsilon_t$, the operation $[A]_{lower}$ 
zeros the upper triangular part  (above the main diagonal) of a squared matrix and
$\text{diag}( \frac{1}{L_{1 1}}, \ldots, \frac{1}{L_{n n}})$ is a diagonal matrix with elements $1/L_{ii}$. The first case of this gradient, i.e.\ when 
 $\log \pi(y_t) < \log \pi(x_t)$, has a very similar structure with the stochastic reparametrisation gradient when fitting 
 a variational Gaussian approximation \citep{Kingma2014,Rezende2014,Titsias2014_doubly}
 with the difference that here  we centre the corresponding approximation, i.e.\ 
 the proposal $q_L(y_t | x_t)$,  at the current state $x_t$ instead of having a global variational mean parameter. 
 Interestingly, this first case when  MCMC rejects many samples (or even it gets stuck at the same value for long time)
 is when learning can be faster since the term $\nabla_{y_t} \log \pi(y_t) \times \epsilon_t^\top$ transfers information about the curvature of the 
 target to the covariance of the proposal. When we start getting high acceptance rates the second case, i.e.\ $\log \pi(y_t) \geq \log \pi(x_t)$, will often be activated 
 so that the gradient will often reduce to only having the term $\beta \text{diag}(\frac{1}{L_{1 1}}, \ldots, \frac{1}{L_{n n}})$ that encourages the entropy of the proposal to become large.  The ability to learn from rejections is in sharp contrast with the traditional non 
 gradient-based adaptive MCMC methods which can become very slow 
  when MCMC has high rejection rates.  This is because these methods typically learn from the history of state vectors $x_t$  
  by ignoring the information from the rejected states. 
 The  algorithm for learning the full random walk Gaussian follows precisely the general structure of Algorithm 1. For the average acceptance rate 
 $\alpha_*$ we use the value $1/4$.   
  
\subsection{Fitting a full covariance MALA proposal \label{sec:gadMALA} }

Here, we specialise to a full covariance, also called preconditioned, MALA of the form
$q_{L}(y  | x ) = \mathcal{N}(y | x + (1 / 2) L L^\top \nabla_{x} \log \pi (x),  L L^\top )$ where the covariance matrix 
is parametrised by the Cholesky factor $L$.  
Again this distribution is reparametrisable according to $y \equiv \mathcal{T}_L (x, \epsilon) = x + (1 / 2) L L^\top \nabla \log \pi(x)  + L \epsilon$, $\epsilon \sim  \mathcal{N}(\epsilon | 0, I)$. 
 At the $t$-th iteration when the state is $x_t$ the reparametrised lower bound simplifies significantly and reduces to,   
 {\small 
 \begin{align}
\mathcal{F}_{L}(x_t) 
 & = \int \mathcal{N}(\epsilon | 0, I) \text{min} \Bigl\{
 0, \log  \pi  \left(x_t +  (1/2) L L^\top \nabla \log \pi(x_t)  + L \epsilon \right)   - \log \pi(x_t) 
 \Bigr. \nonumber \\
 \Bigl. 
& - \frac{1}{2} \left( ||  (1/2) L^\top [ \nabla \log \pi(x_t)  + \nabla \log \pi(y) ]  + \epsilon  ||^2 -  || \epsilon ||^2 \right)
  \Bigr\}
  d \epsilon  +  \beta \sum_{i=1}^n \log L_{i i} + \text{const},  \nonumber
 \end{align}  
 }
 where $|| \cdot ||$ denotes Euclidean norm and in the term $\nabla \log \pi(y)$, $y = x_t +  (1/2) L L^\top \nabla \log \pi(x_t)  + L \epsilon$. 
 Then, based on the proposed state $y_t =  \mathcal{T}_L (x_t, \epsilon_t) $  we can obtain the unbiased gradient estimate $\nabla \mathcal{F}_{L}(x_t, \epsilon_t)$ similarly  to the previous section.  In general, such an estimate can be very expensive because the existence of $L$ inside  
 $\nabla \log \pi(y_t)$ means that we need to compute the matrix of second derivatives or Hessian $\nabla \nabla \log \pi(y_t)$. 
 We have found that an alternative procedure which stops the gradient inside $\nabla \log \pi(y_t)$ (i.e.\  it views $\nabla \log \pi(y_t)$ as 
 a constant w.r.t.\ $L$) has small bias and works well in practice. In fact, as we will show in the experiments this approximation not only is 
 computationally much faster but remarkably also it leads to better adaptation compared to the exact Hessian procedure, presumably 
 because by not accounting for the gradient inside $\nabla \log \pi(y_t)$ reduces the variance.  
 Furthermore,  the expression of the gradient w.r.t.\ $L$ used by this fast approximation 
 can be computed very efficiently with a single $O(n^2)$ operation (an outer vector product; see Supplement), while each 
 iteration of the algorithm requires overall at most four $O(n^2)$ operations. 
 For these gradient-based adaptive MALA schemes, $\beta$ in Algorithm 1 is adapted to 
 obtain an average acceptance rate roughly $\alpha_*=0.55$.


\section{Related Work}
\label{sec:related}

Connection of our method with traditional adaptive MCMC methods has been discussed in Section \ref{sec:introduction}. 
Here, we analyse additional related works that make use of gradient-based optimisation 
and specialised objective functions or algorithms to train MCMC proposal distributions.   

The work in \cite{levy2018generalizing} proposed a criterion to tune MCMC proposals 
based on maximising a modified version of the expected squared jumped distance, 
$\int q_{\theta}(y | x_t) ||y - x_t||^2 \alpha(x_t, y; \theta) d y$, previously considered in \cite{pasarica2010adaptively}. 
Specifically, the authors in \cite{levy2018generalizing} firstly observe that the expected squared jumped distance 
 may not encourage mixing across all dimensions of $x$\footnote{Because 
 the additive form of  $||y - x_t||^2 = \sum_i (y_i - x_{t i})^2$ implies that even when some dimensions
 might not be moving at all (the corresponding distance terms are zero), 
the overall sum can still be large.} and then try to resolve this  
by including a reciprocal term (see Section 4.2 in their paper). The  
generalised speed measure proposed in this paper is rather different from such criteria 
since it encourages joint exploration of all dimensions of $x$ by applying maximum entropy 
 regularisation, which by construction penalises "dimensions that do not move" since the entropy becomes minus infinity in such cases. 
Another important difference is that in our method the optimisation 
is performed in the log space by propagating gradients through the logarithm of the 
M-H acceptance probability, i.e.\  through  $\log \alpha(x_t, y; \theta)$ and not through  $\alpha(x_t, y; \theta)$. This is exactly 
analogous to other numerically stable objectives such as variational lower bounds and log likelihoods, and as those   
our method leads to numerically stable optimisation for arbitrarily large dimensionality of $x$
and complex targets $\pi(x)$. 
     
In another related work, the authors in  \cite{neklyudov2018metropolis} considered minimising the 
KL divergence  $\text{KL}[\pi(x_t) q_{\theta}(y_t | x_t) || \pi(y_t) q_{\theta} (x_t | y_t) ]$.
However, this loss for standard proposal schemes, such as RWM and MALA, 
leads  to degenerate deterministic solutions where $q_{\theta}(y_t | x_t)$ collapses to a delta function. 
Therefore,  \cite{neklyudov2018metropolis}  maximised this objective for the independent M-H sampler
where the collapsing problem does not occur. The entropy regularised objective we introduced is different 
and  it can adapt arbitrary MCMC proposal distributions, and not just the independent M-H sampler. 

There has been also work to learn flexible MCMC proposals using neural  
networks \citep{song2017nice, levy2018generalizing, habib2018auxiliary, Salimans2015}. 
For instance,  \cite{song2017nice} use volume preserving flows and 
an adversarial objective,  \cite{levy2018generalizing} use the modified 
expected jumped distance,  discussed earlier, to learn neural network-based 
extensions of HMC, while \cite{habib2018auxiliary, Salimans2015} use auxiliary variational inference. 
The need to train neural networks can add a significant computational cost,  
and from the end-user point of view these neural adaptive samplers might be hard to tune especially in high dimensions. 
Notice that
the generalised speed measure we proposed in this paper could possibly 
be used to train neural adaptive samplers as well. However, to really 
obtain practical algorithms we need to ensure that training has small cost that does not 
overwhelm the possible benefits in terms of effective sample size. 

Finally, the generalised speed measure that is based on  
entropy regularisation shares similarities with other used objectives for learning 
probability distributions, such as in variational Bayesian inference, where the variational lower bound 
includes an entropy term \citep{Jordan1999learn,Bishop2006} and  reinforcement learning (RL) where  
maximum-entropy regularised policy gradients are able to 
estimate more explorative policies \citep{schulman15, mniha16}. Further discussion on 
the resemblance  of our algorithm with RL is given in the Supplement.

\vspace{-2mm} 

\section{Experiments}
\label{sec:experiments}

\vspace{-2mm} 

We test the gradient-based adaptive MCMC methods in several simulated and real data. 
We investigate the performance of two instances of the framework: 
 the gradient-based adaptive random walk  (gadRWM) 
detailed in Section \ref{sec:gadRWM}  and  the corresponding MALA (gadMALA)
detailed in Section \ref{sec:gadMALA}. For gadMALA we consider 
the exact reparametrisation  method that requires the evaluation of the Hessian (gadMALAe)
 and the fast approximate variant (gadMALAf).     
These schemes are compared against: (i) standard random walk Metropolis (RWM)
with proposal $\mathcal{N}(y|x, \sigma^2 I)$, (ii) an adaptive MCMC 
(AM) that fits a proposal of the form $\mathcal{N}(y|x, \Sigma)$ (we consider 
a computational efficient version based on updating the Cholesky factor of
 $\Sigma$; see Supplement), (iii) a  standard MALA proposal  $\mathcal{N}(y|x + (1/2) \sigma^2 \nabla \log \pi(x), \sigma^2 I)$, 
(iv) an Hamiltonian Monte Carlo (HMC) with a fixed number of leap frog steps (either 5, or 10, or 20) 
(v) and the state of the art no-U-turn sampler (NUTS) \citep{hoffman2014no} which arguably is the most efficient adaptive HMC method 
 that automatically determines the number of leap frog steps.  
 We provide our own MALTAB implementation\footnote{{\tt https://github.com/mtitsias/gadMCMC.}} of all algorithms, apart from NUTS for which we consider a publicly 
 available implementation.  
 
  \vspace{-1.5mm} 
  
 \subsection{Illustrative experiments}
 
 \vspace{-2mm} 
 
\begin{figure*}[!htb]
\centering
\begin{tabular}{cccc}
{\includegraphics[scale=0.17]
{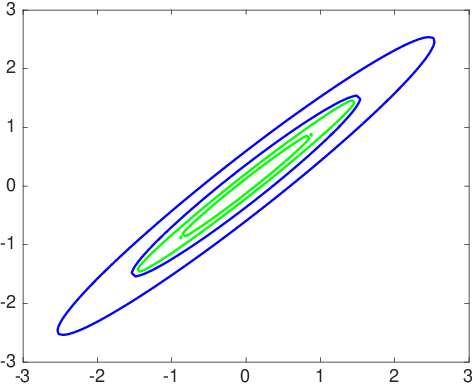}} &
{\includegraphics[scale=0.17]
{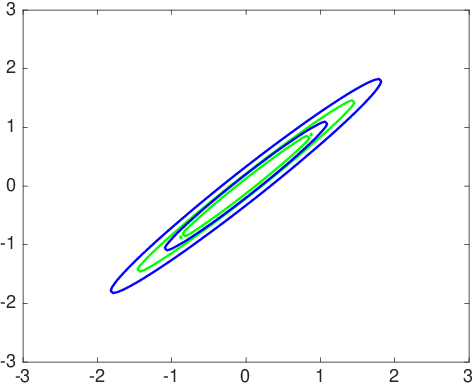}}  &
{\includegraphics[scale=0.17]
{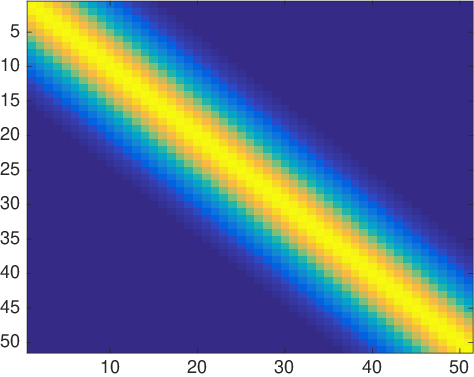}} &
{\includegraphics[scale=0.17]
{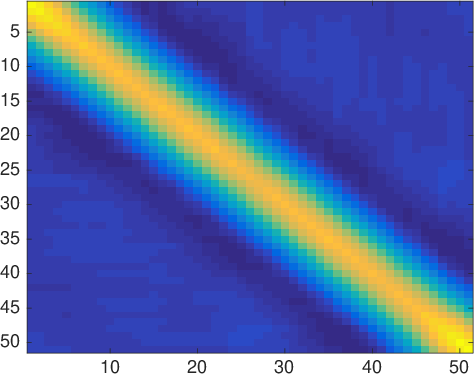}} 
\end{tabular}
\vspace{-2.5mm}
\caption{
The green contours in the first two panels (from left to right) show the 2-D Gaussian target, while the 
blue contours show the learned covariance, $L L^\top$, after adapting for $2 \times 10^4$
iterations using gadRWM and targeting acceptance rates $\alpha_*=0.25$ and $\alpha_*=0.4$, respectively.  
For $\alpha_*=0.25$ the adapted blue contours show that the proposal matches the shape of the target but it has higher entropy/variance  
and  the hyperparameter $\beta$ obtained the value $7.4$. For $\alpha_* = 0.4$ the blue contours shrink a bit and  $\beta$ is 
reduced to $2.2$ (since higher acceptance rate requires smaller entropy).  The third panel shows the exact $51 \times 51$ covariance 
matrix and the last panel shows the adapted one, after running our most efficient 
gadMALAf scheme for $2 \times 10^5$ iterations.  In both experiments $L$ was initialised to diagonal matrix with $0.1/\sqrt{n}$ in the diagonal.} 
\label{fig:illustate}
\end{figure*}

 To visually illustrate the gradient-based adaptive samplers we consider a correlated 2-D Gaussian    
 target with covariance matrix $\Sigma = [1 \ 0.99; 0.99 \ 1]$ and a 51-dimensional Gaussian target obtained 
 by evaluating the squared exponential kernel plus small 
 white noise, i.e.\ $k(x_i,x_j) = \exp\{ - \frac{1}{2} \frac{(x_i - x_j)^2}{0.16} \} + 0.01\delta_{i,j}$,
 on the regular grid $[0,4]$.  The first two panels in Figure \ref{fig:illustate} show the true covariance together  with 
  the adapted covariances obtained by gadRWM for two different settings of the average acceptance rate $\alpha_*$ 
  in Algorithm 1,  which illustrates also the adaptation of the entropy-regularisation hyperparameter $\beta$ that is learnt 
 to obtain a certain $\alpha_*$.  The remaining two plots illustrate the ability to learn a highly correlated $51$-dimensional covariance
 matrix (with eigenvalues ranging from $0.01$ to $12.07$)  by applying our most advanced gadMALAf scheme.  
 
 \vspace{-1.5mm} 
 
\subsection{Quantitative results}
 
 \vspace{-2mm} 
 
 Here,  we compare all algorithms in some standard benchmark problems, such as 
 Bayesian logistic regression, and report effective sample size (ESS) together with other quantitative scores.     


 
\begin{figure*}[!htb]
\centering
\begin{tabular}{cccc}
{\includegraphics[scale=0.17]
{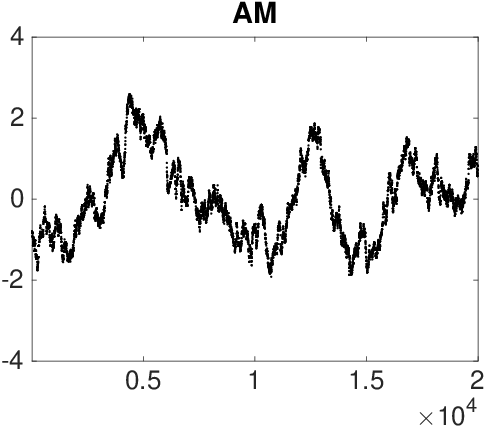}} &
{\includegraphics[scale=0.17]
{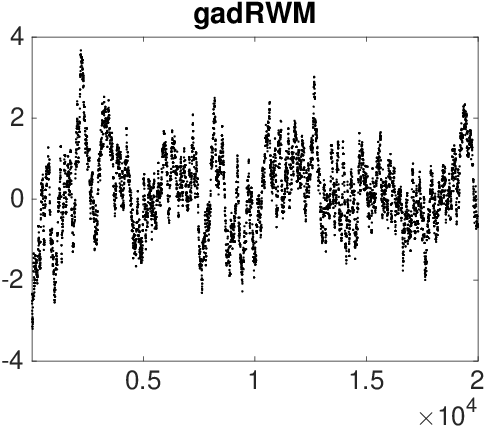}}  &
{\includegraphics[scale=0.17]  
{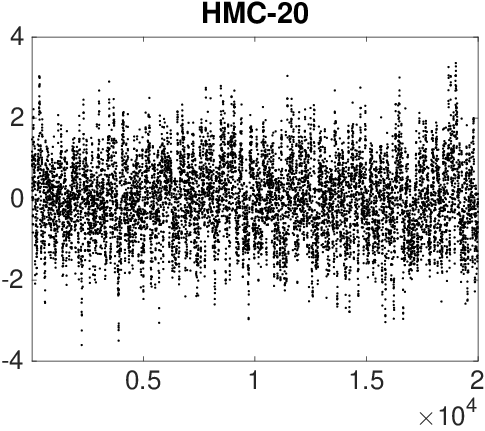}}  &
{\includegraphics[scale=0.17]  
{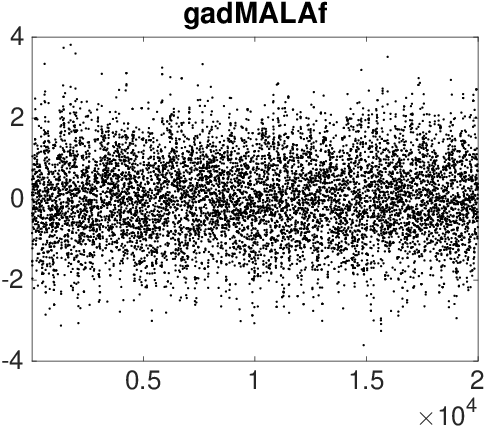}}   \\
{\includegraphics[scale=0.17]
{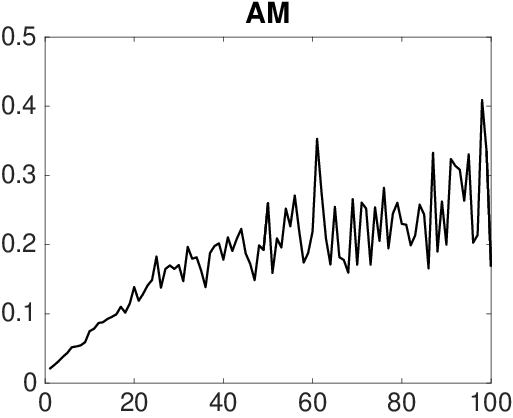}} &
{\includegraphics[scale=0.17]
{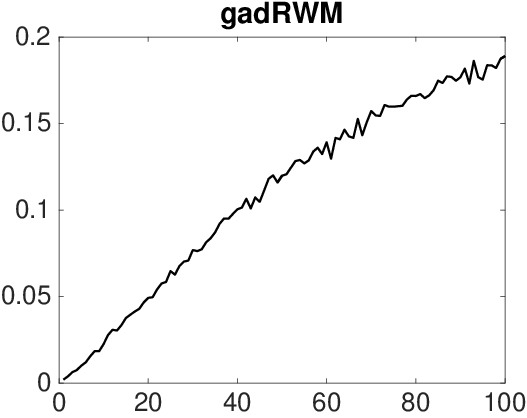}}  &
{\includegraphics[scale=0.17]  
{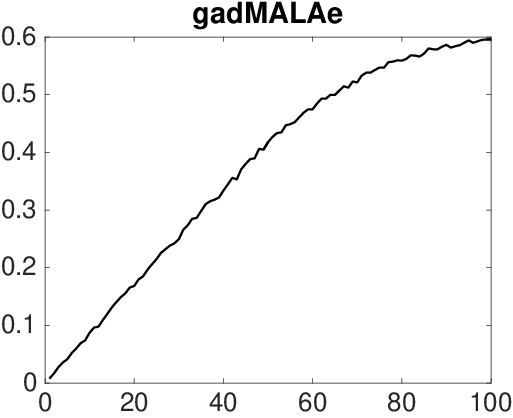}}  &
{\includegraphics[scale=0.17]  
{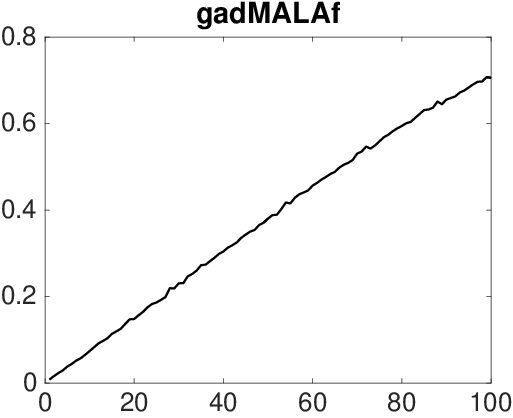}}  
\end{tabular}
\vspace{-2.5mm}
\caption{
Panels in the first row show trace plots, obtained by different schemes, across the last $2 \times 10^4$ sampling 
iterations for the most difficult to sample $x_{100}$ dimension. The panels in the second row show  
the estimated values of the diagonal of $L$ obtained by different adaptive schemes. Notice that the real Gaussian target has 
diagonal covariance matrix $\Sigma = \text{diag}(s_1^2, \ldots, s_{100}^2)$
where $s_i$ are uniform in the range $[0.01,1]$. 
} 
\label{fig:neal}
\end{figure*}

\begin{table}
  \caption{
  Comparison  in Neal's Gaussian example (dimensionality was $n=100$; see panel above)
   and Caravan binary classification  dataset where the latter consists of 5822
  data points (dimensionality  was $n=87$; see panel below).  All numbers are averages across ten repeats where also one-standard deviation is given
for the Min ESS/s score. From the three HMC schemes we report only the best one in each case.  
  }
  \label{table}
  \vspace{-1mm}
  \begin{center}
   \begin{tabular}{lllll}
    \toprule
Method &  Time(s) & Accept Rate &  ESS (Min, Med, Max)  & Min ESS/s (1 st.d.) \\ 
\midrule
{(\em Neal's Gaussian)} \\
gadMALAf  &   8.7  &  0.556  &  (1413.4, 1987.4, 2580.8)  &  {\bf 161.70} (15.07)\\ 
gadMALAe  &   10.0  &  0.541  &  (922.2, 2006.3, 2691.1)  &  92.34 (7.11)\\ 
gadRWM  &   7.0  &  0.254  &  (27.5, 66.9, 126.9)  &  3.95 (0.66)\\ 
AM  &   2.3  &  0.257  &  (8.7, 48.6, 829.1)  &  3.71 (0.87)\\ 
RWM  &   2.2  &  0.261  &  (2.9, 8.4, 2547.6)  &  1.31 (0.06)\\ 
MALA  &   3.1  &  0.530  &  (2.9, 10.0, 12489.2)  &  0.95 (0.03)\\ 
HMC-20  &   49.7  &  0.694  &  (306.1, 1537.8, 19732.4)  &  6.17 (3.35)\\ 
NUTS  &   360.5  &  >0.7  &  (18479.6, 20000.0, 20000.0)  &  51.28 (1.64)\\ 
%
\midrule
{(\em Caravan)} \\
gadMALAf  &   23.1  &  0.621  &  (228.1, 750.3, 1114.7)  &  {\bf 9.94} (2.64)\\ 
gadMALAe  &   95.1  &  0.494  &  (66.6, 508.3, 1442.7)  &  0.70 (0.16)\\ 
gadRWM  &   22.6  &  0.234  &  (5.3, 34.3, 104.5)  &  0.23 (0.06)\\ 
AM  &   20.0  &  0.257  &  (3.2, 11.8, 62.5)  &  0.16 (0.01)\\ 
RWM  &   15.3  &  0.242  &  (3.0, 9.3, 52.5)  &  0.20 (0.03)\\ 
MALA  &   22.8  &  0.543  &  (4.4, 28.3, 326.0)  &  0.19 (0.05)\\ 
HMC-10  &   225.5  &  0.711  &  (248.3, 2415.7, 19778.7)  &  1.10 (0.12)\\ 
NUTS  &   1412.1  &  >0.7  &  (7469.5, 20000.0, 20000.0)  &  5.29 (0.38)\\ 
%
\bottomrule
  \end{tabular}
  \end{center}
\end{table}

{\bf Experimental settings.} In all experiments for AM and gradient-based adaptive schemes 
 the Cholesky factor $L$ was  initialised to a diagonal matrix with values $0.1/\sqrt{n}$ in the diagonal where $n$ is 
 the dimensionality of $x$.  For the AM algorithm the learning rate was set  to
  $0.001/(1 + t/T)$ where $t$ is the number of iterations and $T$ (the value $4000$ was used in all experiments) 
  serves to keep the learning rate nearly constant for the 
 first $T$ training iterations.  For the gradient-based adaptive algorithms 
we use RMSprop (see Section \ref{sec:maximisespeed}) where $\eta$ was set to $0.00005$ 
for gadRWM and  to $0.00015$ for the gadMALA schemes. 
 NUTS uses its own fully automatic adaptive 
 procedure that determines both the step size and the number of leap frog steps  
 \citep{hoffman2014no}.  For all experiments and algorithms (apart from NUTS) 
 we consider $2 \times 10^4$ burn-in iterations and $2 \times 10^4$ iterations for 
collecting samples. This adaptation of $L$ or $\sigma^2$ takes place only during the burn-in iterations and then it stops,
i.e.\ at collection of samples stage these parameters are kept fixed. For NUTS, which has its own 
internal tuning procedure, $500$ burn-in iterations are sufficient 
before collecting  $2 \times 10^4$ samples. The computational 
time for all algorithms reported in the tables corresponds to the overall running 
time, i.e.\ the time for performing jointly all burn-in and collection of samples iterations. 


\noindent  {\bf Neal's Gaussian target.}
We first consider an example used in \citep{neal2010} 
where the target is a zero-mean multivariate Gaussian with diagonal covariance matrix 
$\Sigma = \text{diag}(s^2_1,\ldots, s_{100}^2)$ where the stds $s_i$ 
take values in the linear grid $0.01,0.02, \dots,1$.  This is a challenging example because
 the different scaling of the dimensions means that the schemes that use an isotropic step 
$\sigma^2$ will be adapted to the smallest dimension $x_1$ while the chain at the 
higher dimensions, such as $x_{100}$, will be moving slowly exhibiting high autocorrelation and small 
effective sample size. The first row of Figure \ref{fig:neal} shows the trace plot across
iterations of the dimension $x_{100}$  for some of the adaptive schemes 
including an HMC scheme that uses $20$ leap frog steps.  Clearly, the gradient-based adaptive methods
show much smaller autocorrelation that AM.  This is because they achieve a more 
efficient adaptation of the Cholesky factor $L$ which ideally should become proportional to a diagonal matrix 
with the linear grid $0.01,0.02, \dots,1$ in the main diagonal. 
The second row of Figure \ref{fig:neal} shows the diagonal elements of $L$ from which we can observe that 
all gradient-based schemes lead to more accurate adaptation with gadMALAf being the most accurate.  
      
Furthermore, Table \ref{table} provides quantitative results such as minimum, median  and maximum 
ESS computed across all dimensions of the state vector $x$, running times and
an overall efficiency score Min ESS/s (i.e.\ ESS for the slowest moving component of $x$ divided by running time -- last column in the Table) 
which allows to rank the different algorithms. All results  are averages after 
repeating the simulations $10$ times under different random initialisations.   
From the table it is  clear that the gadMALA algorithms give the best performance 
with gadMALAf being overall the most effective. 

\noindent {\bf Bayesian logistic regression.} We consider binary classification
where given a set of training examples $\{y_i, s_i\}_{i=1}^n$ 
we assume a logistic regression likelihood 
$p(y |w, s) = \sum_{i=1}^n y_i \log \sigma(s_i) + (1 - y_i) \log (1 - \sigma(s_i))$,
where $\sigma(s_i) = 1/(1 + \exp(-w^\top s_i ))$, $s_i$ is the input vector and $w$ 
the parameters. We place a Gaussian prior on $w$ and we wish to sample from the posterior 
distribution over $w$.  We considered six binary classification datasets (Australian Credit, Heart, Pima Indian, Ripley, German Credit and Caravan) with a number of examples ranging from $n=250$ to $n=5822$ and dimensionality of the state/parameter vector ranging from $3$ to $87$.
Table  \ref{table} shows results for the most challenging  
Caravan dataset where the dimensionality of $w$ is $87$, while the remaining  
five tables are given in the Supplement.  From all tables we observe that the gadMALAf is the 
most effective and it outperforms all other methods. While NUTS has always very high ESS is still 
outperformed by gadMALAf because of the high computational cost, i.e. NUTS 
might need to use a very large number of leap frog steps (each requiring  
re-evaluating the gradient of the log target) per iteration. 
Further results, including a higher $785$-dimensional example on MNIST, 
are given in the Supplement.

\vspace{-3mm}

\section{Conclusion}
\label{sec:conclusion}
\vspace{-2mm}

We have presented a new framework for gradient-based adaptive MCMC
that makes use of an  entropy-regularised objective function that generalises 
the concept of speed measure. We have applied this method for learning RWM
and MALA proposals with full covariance matrices. 
      
Some topics for future research are the following. Firstly, 
to deal with very high dimensional spaces it would be useful to consider 
low rank parametrisations of the covariance matrices in RWM and MALA proposals. 
Secondly, it would be  interesting 
to investigate whether our method can be used to tune the so-called mass matrix in 
HMC samplers.  However, in order for this to lead to 
practical and scalable algorithms we have to come up with schemes  that avoid the 
computation of the Hessian, as we successfully have done for  MALA.  Finally,  
in order to reduce the variance of the stochastic gradients and speed up further the adaptation, 
especially in high dimensions,  our framework could be possibly combined with parallel computing as used for instance 
in deep reinforcement learning \cite{espeholt18a}. 


\bibliographystyle{plain}

\bibliography{my_bibliography,fjrrLibrary}

\appendix

\clearpage

\begin{center}
{\Large \bf Supplement: \\ 
Gradient-based Adaptive Markov Chain Monte Carlo}
\end{center}

\section{Gradient-based adaptive MCMC as Reinforcement Learning }

A MDP is  a tuple $(\mathcal{X}, \mathcal{Y}, \mathcal{P}, \mathcal{R})$ 
where $\mathcal{X}$  is the state space, $\mathcal{Y}$ is the action space, 
$ \mathcal{P}$ is the transition distribution with density $p(x_{t+1} | x_t, y_t)$ 
that describes how the next state $x_{t+1}$ is generated given that
 currently we are at state $x_t$ and we take action $y_t$. Further, the reward function
 $R(x_t, y_t)$ provides some instantaneous  or local signal about 
 how good the  action $y_t$ was when being at $x_t$.  Furthermore, in a MDP 
 we have also a policy $\pi(y_t | x_t)$ which is a distribution over actions given states and it 
 fully describes the behaviour of the agent.  Given that we start at $x_0$
 we wish to specify the policy so that to maximise 
future reward, such as 
the expected accumulated discounted  reward 
$$
\mathbbm{E}_{\pi}\left[  \sum_{t=0}^{\infty } \gamma^t R(x_t, y_t) \right], \ \ \gamma \in [0,1]. 
$$
 Suppose now a MCMC procedure targeting $\pi(x)$, 
 where $x \in \mathcal{X}$  is the state vector.
 Consider a  proposal
 distribution $q_{\theta}(y|x)$, such that the standard Metropolis-Hastings 
  algorithm accepts each proposed state $y_t \sim q_{\theta}(y_t | x_t)$ with probability 
 \begin{equation}
   \alpha(x_t,y_t; \theta) = \text{min}\left(1, \frac{\pi(y_t)}{\pi(x_t)}
   \frac{ q_{\theta}(x_t | y_t) }{q_{\theta}(y_t | x_t)} \right),
\label{eq:acceptanceProb}
 \end{equation}
 so that $x_{t+1} = y_t$, while if the proposal is rejected, $x_{t+1} = x_t$. To reformulate 
MCMC  as an MDP we make the following correspondences. Firstly,  
 both the state $x_t$ and the action $y_t$ will live in the same space which will be the state space 
$\mathcal{X}$ of the target distribution. The MCMC proposal $q_{\theta}(y_t | x_t)$ 
will correspond to the policy
$\pi(y_t | x_t)$, while the environmental transition dynamics  will be stochastic 
and given by the two-component mixture,
$$
p(x_{t+1} | x_t, y_t)  = \alpha(x_t,y_t; \theta)  \delta_{x_{t+1}, y_t} + (1 - \alpha(x_t,y_t; \theta) ) \delta_{x_{t+1}, x_t}, 
$$
where $\delta_{x,y}$ denotes the delta function. This transition density simply says that the new 
state $x_{t+1}$ with probability  $\alpha(x_t,y_t;\theta)$ will be equal to the proposed action $y_t$, while 
with the remaining probability will be set to the previous state, i.e.\ $x_{t+1} = x_t$.  
Notice that the standard MCMC transition kernel $K_{\theta}(x_t, x_{t+1})$
is obtained by integrating out the action $y_t$, i.e.\  
\begin{align}
K_{\theta}(x_t, x_{t+1})  & = \int p(x_{t+1} | x_t, y_t) q_{\theta}(y_t | x_t) d y_t  \nonumber \\
& = 
\alpha(x_t, x_{t+1})  q(x_{t+1} | x_t) +  \left( 1 - \int \alpha(x_t,y_t) q_{\theta}(y_t | x_t) d y_t \right) \delta_{x_{t+1}, x_t}. 
\end{align}
The final ingredient we need to reformulate MCMC as MDP is the reward function $R(x_t, y_t)$. 
The gradient-based adaptive MCMC method  
essentially assumes as reward 
$$
R(y_t, x_t; \theta) = \log \alpha(x_t,y_t; \theta) -  \beta \log q_{\theta}(y_t | x_t),
$$
which is an entropy-regularised reward that promotes high exploration with the entropic 
term  $ - \beta \log q_{\theta}(y_t | x_t)$.  Gradient-based adaptive MCMC essentially at 
each step stochastically maximises the expected reward starting from state $x_t$, i.e.\
 $$
 \int q_{\theta}(y_t  | x_t)  R(y_t, x_t; \theta)  d y_t. 
 $$ 
While the above reformulates MCMC as a reinforcement learning (RL) problem, there are clearly 
also some differences with standard RL problems.  Given that the reward  
$R(y_t, x_t; \theta)$ is very informative (we are not facing the delayed-reward  problem commonly encountered  
in standard RL) gradient-based MCMC sets $\gamma=0$ in order to maximise immediate reward.
Further,  the transition dynamics  $p(x_{t+1} | x_t, y_t)$ are known in MCMC, while this typically is not the case 
in  standard RL. Finally, notice that  the reward $R(y_t, x_t; \theta)$ as  well as 
the transition dynamics  $p(x_{t+1} | x_t, y_t)$ all depend on the parameter $\theta$ and 
the policy $q_{\theta}(y_t|x_t)$,  i.e.\ they depend on the MCMC proposal distribution.

\section{Further details about the algorithms }  

For the standard adaptive MCMC method (AM) we implemented a computational efficient version that 
requires no matrix decompositions (which are expensive due to the $O(n^3)$ scaling) by parametrising 
the proposal as $\mathcal{N}(y|x, L L^\top)$ and updating the Cholesky factor in each iteration 
according to the updates   
$$
\mu \leftarrow \mu + \rho_t (x_{t+1} - \mu), 
$$
$$
L  \leftarrow L + \rho_t   L \left[  
L^{-1} (x_{t+1} - \mu) (x_{t+1} - \mu)^\top L^{-\top} -  I  
\right]_{lower},
$$
where $\mu$ tracks the global mean of the state vector. Further details about this 
scheme can be found in Section 5.1.1 in   \cite{Andrieu:2008}.

For our most efficient gadMALAf scheme 
 the stochastic gradient in each iteration is 
 {\small 
 \begin{align}
\nabla \mathcal{F}_{L}(x_t, \epsilon_t) 
 & = \nabla_L \text{min} \Bigl\{
 0, \log  \pi  \left(x_t +  (1/2) L L^\top \nabla \log \pi(x_t)  + L \epsilon_t \right)   - \log \pi(x_t) 
 \Bigr. \nonumber \\
 \Bigl. 
& - \frac{1}{2} \left( ||  (1/2) L^\top [ \nabla \log \pi(x_t)  + \nabla \log \pi(y_t) ]  + \epsilon_t  ||^2 -  || \epsilon_t ||^2 \right)
  \Bigr\}
    +  \beta \nabla_L \sum_{i=1}^n \log L_{i i},  \nonumber
 \end{align}  
 }
 where, as discussed in the main paper, $\nabla \log \pi(y_t)$ is taken as constant w.r.t.\ $L$. 
 Then the gradient of the M-H log ratio  (when this log ratio is negative, since otherwise its gradient is zero) simplifies as 
   {\small 
 \begin{align}
& \nabla_L   \log  \pi  \left(x_t +  (1/2) L L^\top \nabla \log \pi(x_t)  + L \epsilon_t \right)   
 - \frac{1}{2} \nabla_L  ||  (1/2) L^\top [ \nabla \log \pi(x_t)  + \nabla \log \pi(y_t) ]  + \epsilon_t  ||^2 
  \nonumber \\
&  = 
  - \frac{1}{2}\Bigl(  \nabla \log \pi(x_t)  - \nabla \log \pi(y_t) \Bigr) 
     \left( (1/2) L^\top [ \nabla \log \pi(x_t) - \nabla \log \pi(y_t) ] +  \epsilon_t   \right)^\top \nonumber
 \end{align}  
 }
and then take the lower triangular part. This is just an outer vector product that scales as $O(n^2)$.  
Overall each iteration of the algorithm can be implemented (plus the extra overhead of  
a single gradient evaluation $\nabla \log \pi(y_t)$ of the log target at the proposed state $y_t$) 
by using at most four $O(n^2)$ operations during adaptation 
and exactly two $O(n^2)$ operations after burn-in, as shown in the released code.

\section{Extra results on the Neal's Gaussian target}

Figure \ref{fig:neal} shows trace plot for  the log density of the target 
for all different algorithms.  

\clearpage

\begin{figure}
\centering
\begin{tabular}{ccc}
\includegraphics[width=35mm,height=32mm]
{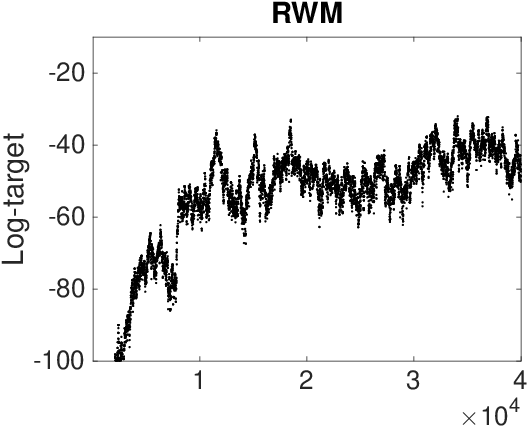} &
\includegraphics[width=35mm,height=32mm]
{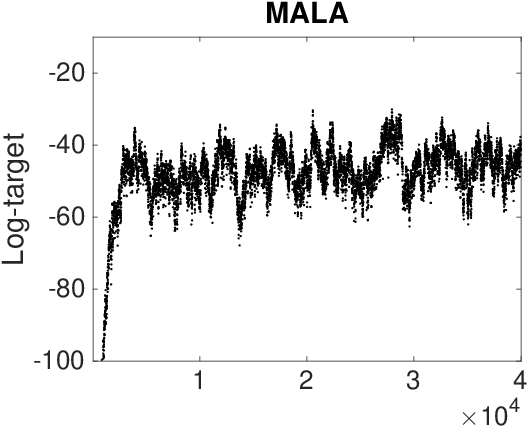} &
\includegraphics[width=35mm,height=32mm]
{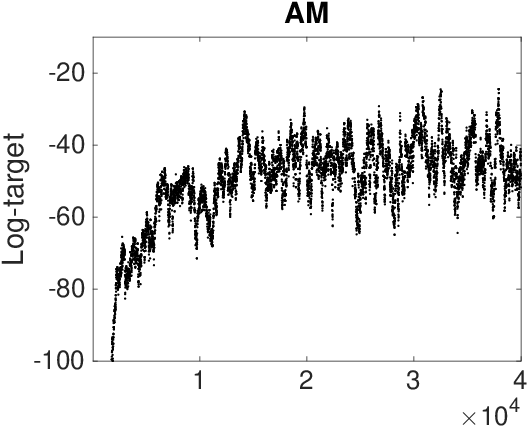} \\
\includegraphics[width=35mm,height=32mm]
{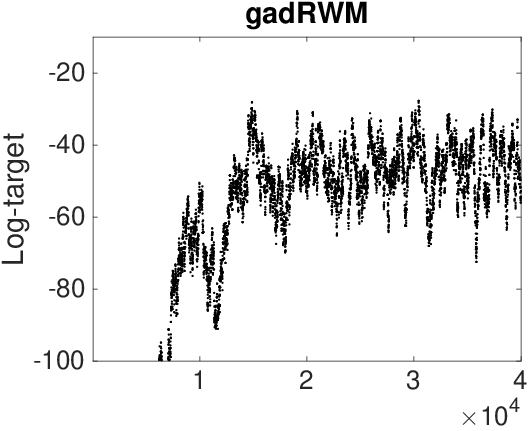} &
\includegraphics[width=35mm,height=32mm]
{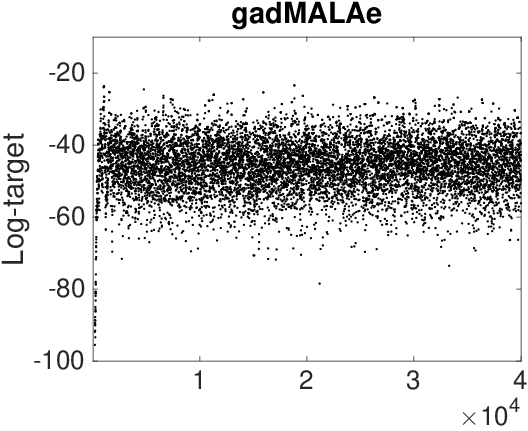} &
\includegraphics[width=35mm,height=32mm]
{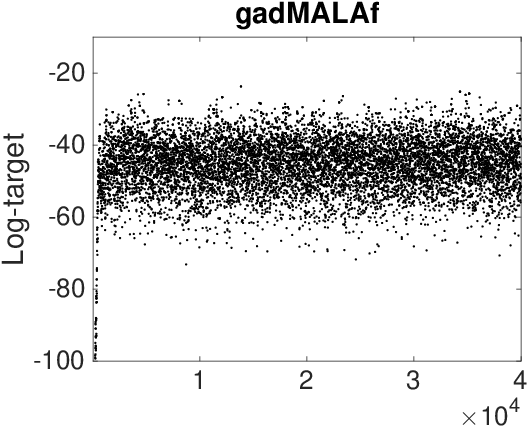} \\
\includegraphics[width=35mm,height=32mm]
{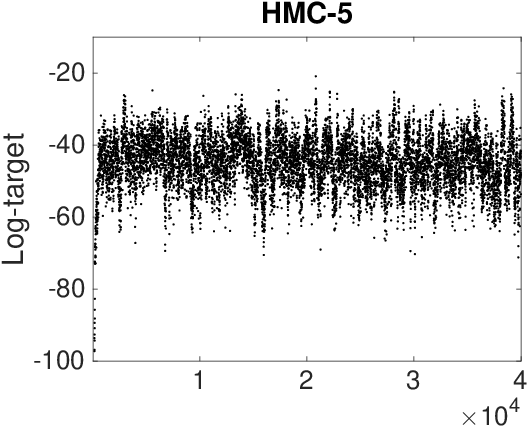} &
\includegraphics[width=35mm,height=32mm]
{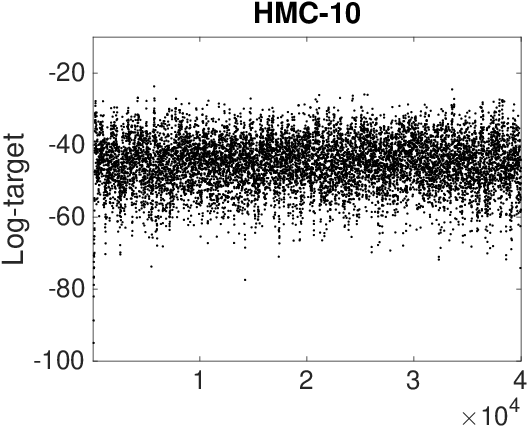} &
\includegraphics[width=35mm,height=32mm]
{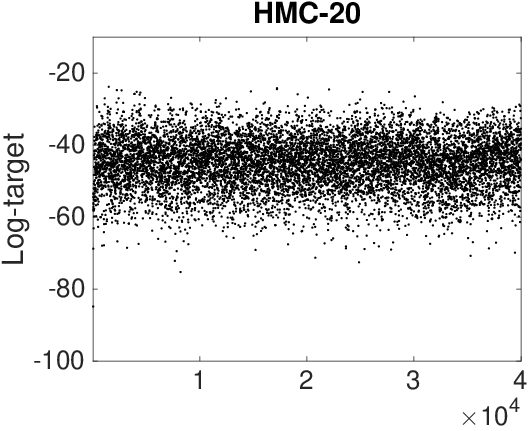} \\
\includegraphics[width=35mm,height=32mm]
{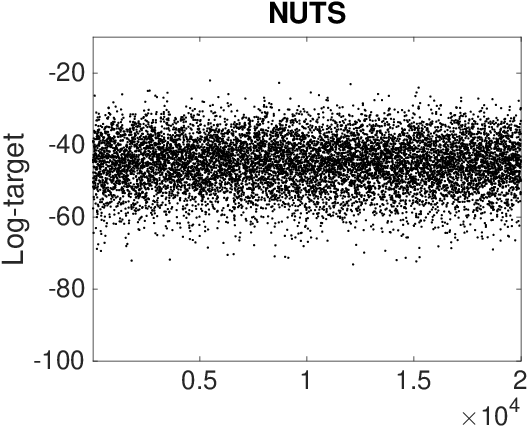} &
\end{tabular}
\caption{The evolution of  the log-target across iterations for all algorithms in Neal's Gaussian example.} 
\label{fig:neal}
\end{figure}

\section{Extra results on the binary classification datasets}

\begin{figure}
\centering
\begin{tabular}{ccc}
\includegraphics[width=36mm,height=32mm]
{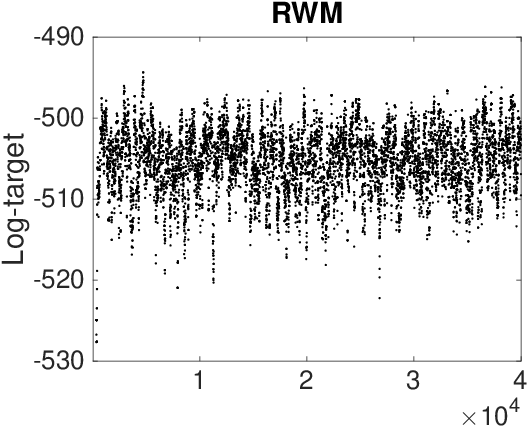} &
\includegraphics[width=36mm,height=32mm]
{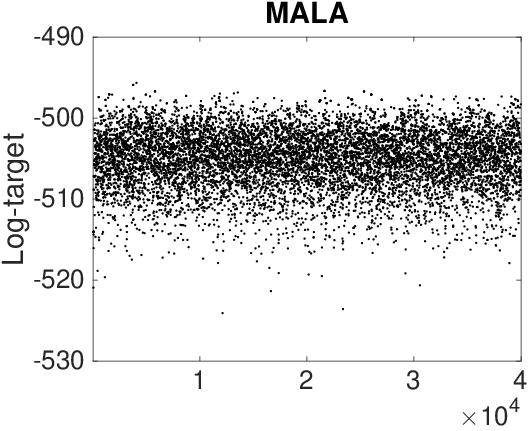} &
\includegraphics[width=36mm,height=32mm]
{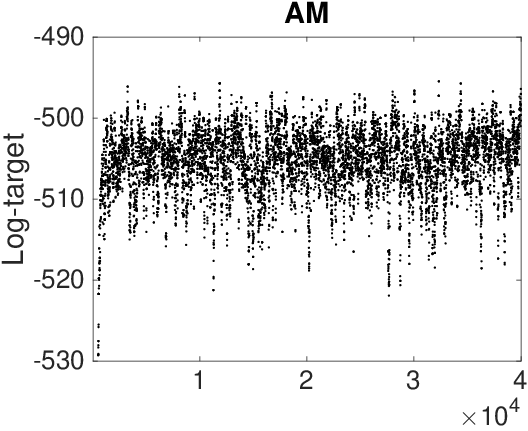} \\
\includegraphics[width=36mm,height=32mm]
{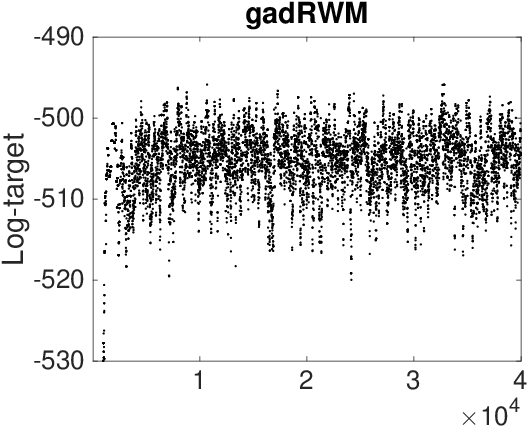} &
\includegraphics[width=36mm,height=32mm]
{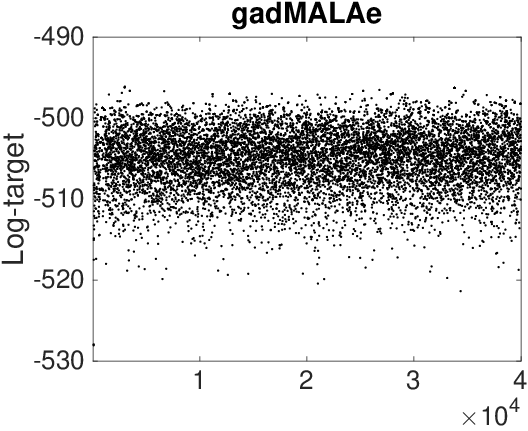} &
\includegraphics[width=36mm,height=32mm]
{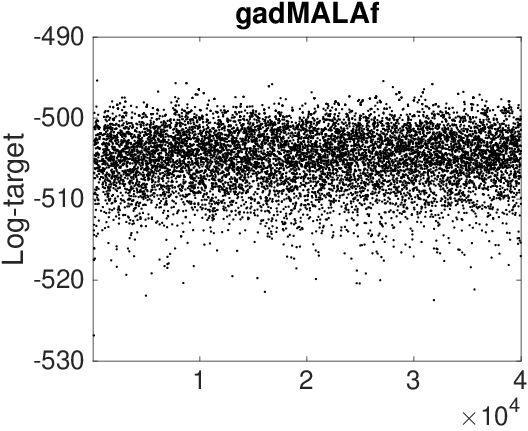} \\
\includegraphics[width=36mm,height=32mm]
{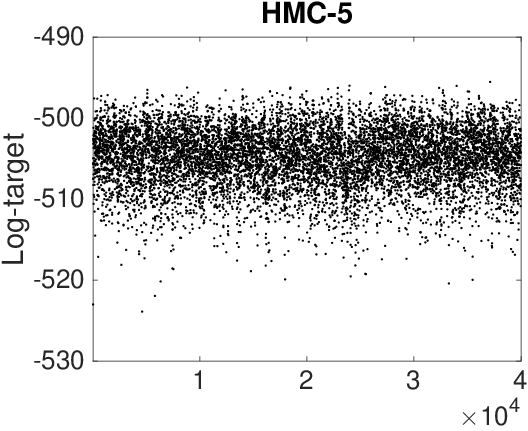} &
\includegraphics[width=36mm,height=32mm]
{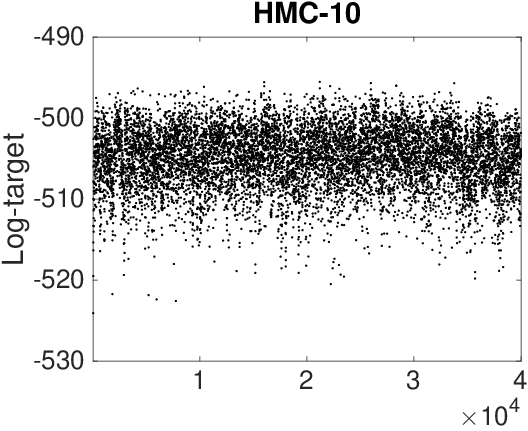} &
\includegraphics[width=36mm,height=32mm]
{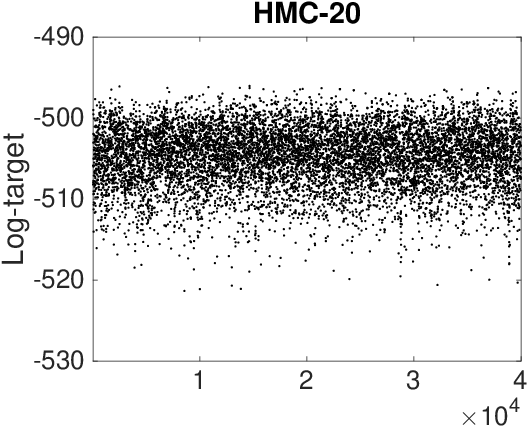} \\
\includegraphics[width=36mm,height=32mm]
{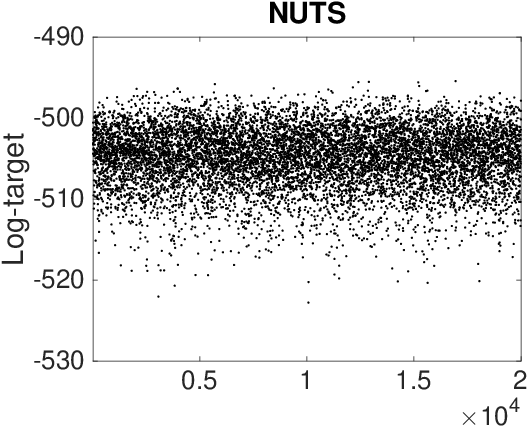} &
\end{tabular}
\caption{The evolution of  the log-target across iterations for all algorithms in German Credit dataset.} 
\label{fig:german}
\end{figure}

\begin{figure}
\centering
\begin{tabular}{ccc}
\includegraphics[width=36mm,height=32mm]
{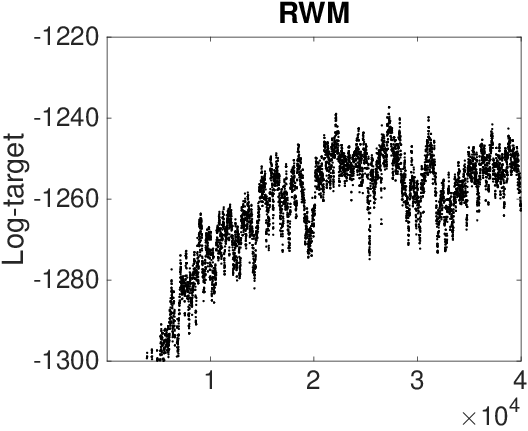} &
\includegraphics[width=36mm,height=32mm]
{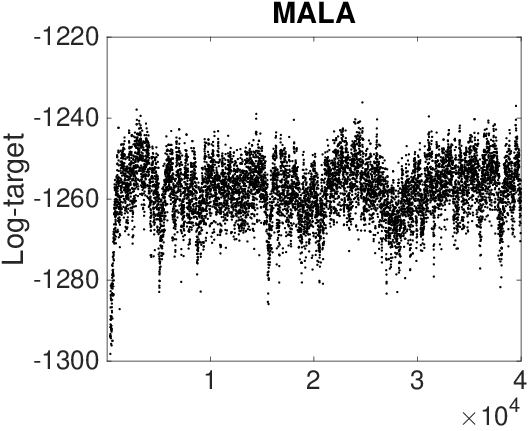} &
\includegraphics[width=36mm,height=32mm]
{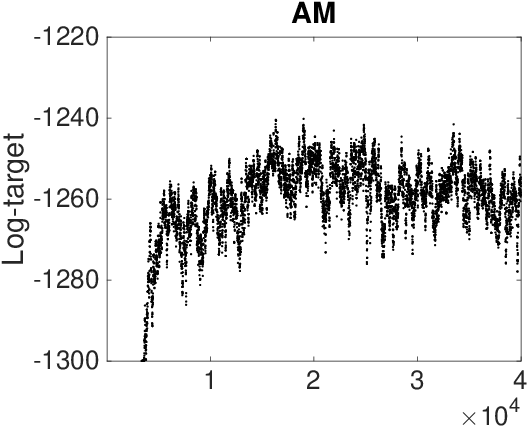} \\
\includegraphics[width=36mm,height=32mm]
{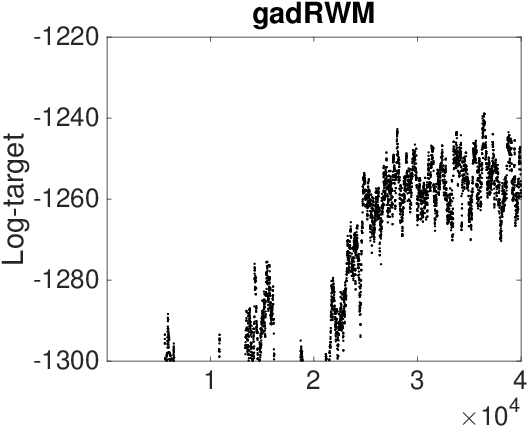} &
\includegraphics[width=36mm,height=32mm]
{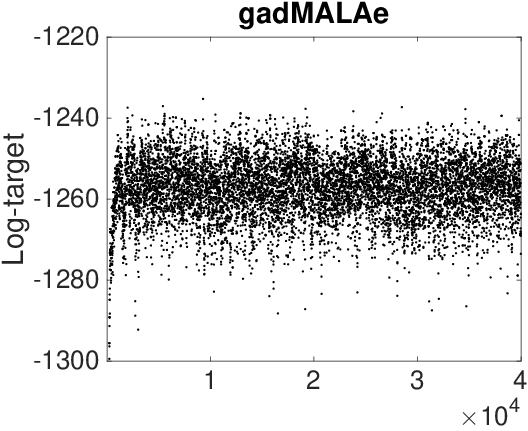} &
\includegraphics[width=36mm,height=32mm]
{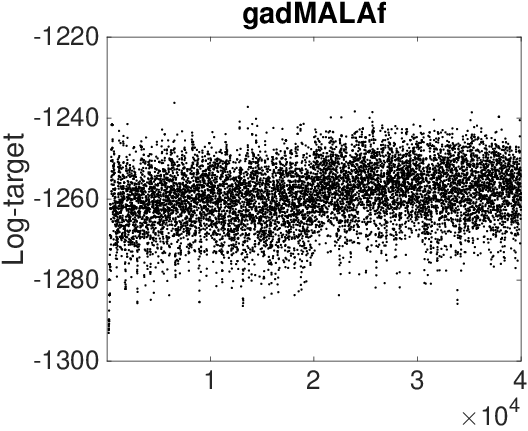} \\
\includegraphics[width=36mm,height=32mm]
{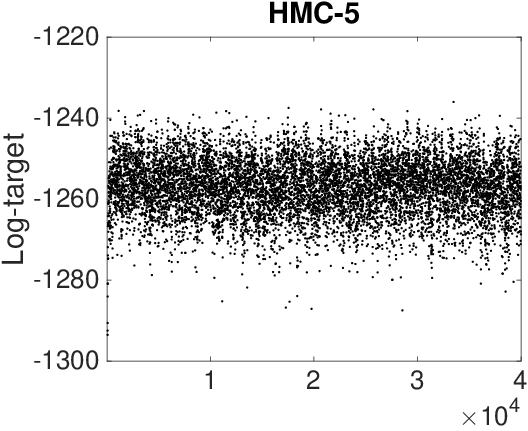} &
\includegraphics[width=36mm,height=32mm]
{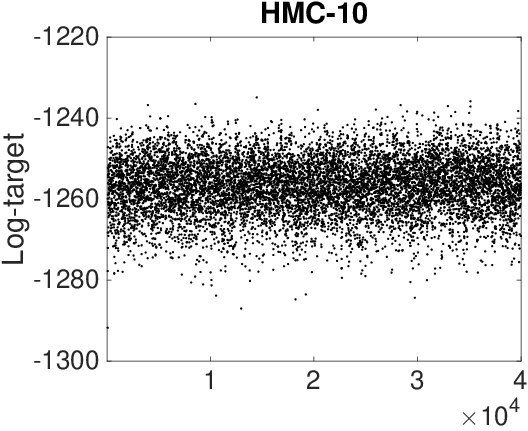} &
\includegraphics[width=36mm,height=32mm]
{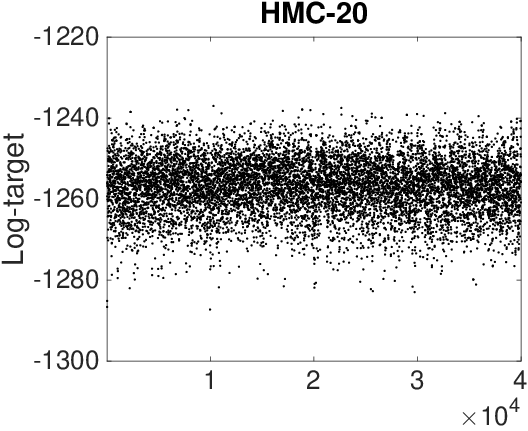} \\
\includegraphics[width=36mm,height=32mm]
{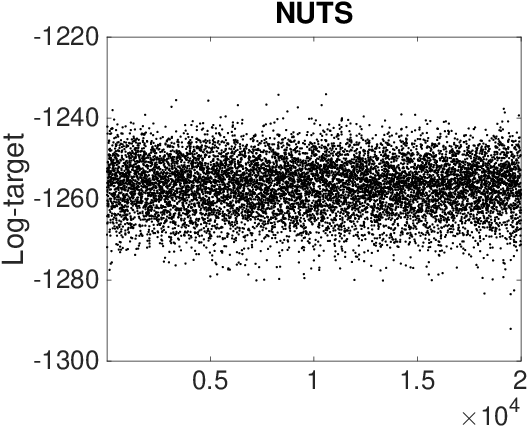} &
\end{tabular}
\caption{The evolution of  the log-target across iterations for all algorithms in Caravan dataset.} 
\label{fig:caravan}
\end{figure}

Tables \ref{table:australian}-\ref{table:german} show the results for the remaining 
five binary classification datasets not reported in the main article. 
Figures \ref{fig:german} and \ref{fig:caravan} show trace plot of log density of the target 
acrorss all different algorithms for the German Credit and Caravan datasets. For the remaining datasets 
the  corresponding plots are similar.  

\begin{table}
  \caption{Comparison of sampling methods in Australian Credit dataset consisted of 690
  data points. The size of the state/parameter vector from which we draw samples was  $n = 15$.}
  \label{table:australian}
  \centering
  \begin{tabular}{lllll}
    \toprule
Method &  Time(s) & Accept Rate  &  ESS (Min, Med, Max)  & Min ESS/s (1 st.d.) \\ 
\midrule
gadMALAf  &   8.1  &  0.569  &  (3485.9, 4262.9, 4784.0)  &  443.97 (76.13)\\ 
gadMALAe  &   13.5  &  0.540  &  (3034.9, 4234.3, 4836.3)  &  227.61 (29.87)\\ 
gadRWM  &   7.7  &  0.253  &  (288.0, 423.0, 515.0)  &  38.68 (9.53)\\ 
AM  &   4.4  &  0.261  &  (310.9, 410.1, 507.2)  &  70.21 (6.23)\\ 
RWM  &   3.4  &  0.252  &  (31.3, 312.6, 495.2)  &  9.16 (3.12)\\ 
MALA  &   7.0  &  0.524  &  (138.4, 2388.1, 3818.8)  &  20.22 (5.14)\\ 
HMC-5  &   37.0  &  0.700  &  (1048.1, 3510.3, 14809.7)  &  28.06 (11.69)\\ 
NUTS  &   41.3  &  >0.7  &  (2995.2, 20000.0, 20000.0)  &  72.86 (7.31)\\ 

  \bottomrule
  \end{tabular}
\end{table}

\begin{table}
  \caption{Comparison of sampling methods in Ripley dataset consisted of 250
  data points. The size of the state/parameter vector from which we draw samples was  $n = 3$.}
  \label{table:ripley}
  \centering
  \begin{tabular}{lllll}
    \toprule
Method &  Time(s) & Accept Rate  &  ESS (Min, Med, Max)  & Min ESS/s (1 st.d.) \\ 
\midrule
gadMALAf  &   3.3  &  0.536  &  (8328.4, 8913.2, 9442.4)  &  2506.04 (143.47)\\ 
gadMALAe  &   4.9  &  0.543  &  (8446.7, 9006.6, 9595.6)  &  1713.44 (44.91)\\ 
gadRWM  &   3.1  &  0.068  &  (638.0, 736.9, 803.2)  &  205.99 (17.85)\\ 
AM  &   3.0  &  0.257  &  (1702.8, 1792.2, 1902.0)  &  570.19 (49.32)\\ 
RWM  &   2.1  &  0.252  &  (1129.2, 1627.8, 1979.8)  &  534.21 (43.54)\\ 
MALA  &   2.8  &  0.542  &  (2976.0, 5683.0, 9726.5)  &  1046.48 (54.86)\\ 
HMC-5  &   14.7  &  0.678  &  (9205.3, 10818.1, 16136.5)  &  626.55 (196.48)\\ 
NUTS  &   7.5  &  >0.7  &  (9436.2, 17463.5, 20000.0)  &  1265.99 (73.01)\\ 

  \bottomrule
  \end{tabular}
\end{table}

\begin{table}
  \caption{Comparison of sampling methods in Pima Indian dataset consisted of 532
  data points. The size of the state/parameter vector from which we draw samples was  $n = 8$.}
  \label{table:pima}
  \centering
  \begin{tabular}{lllll}
    \toprule
Method &  Time(s) & Accept Rate  &  ESS (Min, Med, Max)  & Min ESS/s (1 st.d.) \\ 
\midrule
gadMALAf  &   4.6  &  0.545  &  (5407.6, 5810.3, 6467.6)  &  1176.12 (79.54)\\ 
gadMALAe  &   6.8  &  0.547  &  (5469.6, 5963.6, 6421.1)  &  801.03 (16.07)\\ 
gadRWM  &   4.2  &  0.267  &  (635.6, 760.0, 866.2)  &  150.70 (9.73)\\ 
AM  &   4.1  &  0.273  &  (612.7, 729.1, 854.8)  &  149.18 (10.40)\\ 
RWM  &   3.2  &  0.246  &  (354.6, 496.4, 709.6)  &  111.81 (6.16)\\ 
MALA  &   4.0  &  0.509  &  (1524.9, 2457.2, 3853.6)  &  377.17 (25.80)\\ 
HMC-5  &   20.3  &  0.711  &  (7295.7, 12798.7, 18267.4)  &  359.22 (103.55)\\ 
NUTS  &   15.2  &  >0.7  &  (15343.3, 18606.0, 20000.0)  &  1008.97 (42.33)\\ 

  \bottomrule
  \end{tabular}
\end{table}

\begin{table}
  \caption{Comparison of sampling methods in Heart dataset consisted of 270
  data points. The size of the state/parameter vector from which we draw samples was  $n = 14$.}
  \label{table:heart}
  \centering
  \begin{tabular}{lllll}
    \toprule
Method &  Time(s) & Accept Rate  &  ESS (Min, Med, Max)  & Min ESS/s (1 st.d.) \\ 
\midrule
gadMALAf  &   4.1  &  0.551  &  (3892.9, 4362.7, 4784.2)  &  946.98 (56.10)\\ 
gadMALAe  &   6.4  &  0.560  &  (3832.4, 4372.3, 4845.6)  &  599.51 (30.00)\\ 
gadRWM  &   3.8  &  0.288  &  (342.5, 440.9, 536.1)  &  88.94 (10.29)\\ 
AM  &   3.2  &  0.238  &  (342.5, 425.5, 535.4)  &  106.97 (7.18)\\ 
RWM  &   2.3  &  0.266  &  (196.9, 314.3, 472.7)  &  86.57 (11.33)\\ 
MALA  &   3.5  &  0.530  &  (1429.7, 2310.6, 3260.4)  &  404.96 (18.57)\\ 
HMC-5  &   18.4  &  0.699  &  (1913.2, 5600.3, 11883.0)  &  103.81 (39.38)\\ 
NUTS  &   15.4  &  >0.7  &  (20000.0, 20000.0, 20000.0)  &  1295.13 (15.74)\\ 

  \bottomrule
  \end{tabular}
\end{table}

\begin{table}
  \caption{Comparison of sampling methods in German Credit dataset consisted of 1000
  data points. The size of the state/parameter vector from which we draw samples was  $n = 25$.
  }
  \label{table:german}
  \centering
  \begin{tabular}{lllll}
    \toprule
Method &  Time(s) & Accept Rate  &  ESS (Min, Med, Max)  & Min ESS/s (1 st.d.) \\ 
\midrule
gadMALAf  &   11.0  &  0.560  &  (2734.9, 3414.5, 3928.6)  &  252.91 (37.86)\\ 
gadMALAe  &   22.4  &  0.549  &  (2808.2, 3384.9, 3883.5)  &  126.00 (14.68)\\ 
gadRWM  &   10.4  &  0.248  &  (179.1, 252.5, 323.1)  &  17.92 (4.18)\\ 
AM  &   12.6  &  0.262  &  (121.9, 207.6, 308.0)  &  9.72 (1.25)\\ 
RWM  &   8.4  &  0.233  &  (45.0, 153.8, 298.7)  &  5.48 (1.83)\\ 
MALA  &   9.2  &  0.535  &  (420.1, 1313.2, 2573.7)  &  47.37 (10.22)\\ 
HMC-5  &   43.4  &  0.706  &  (3020.2, 10294.4, 20000.0)  &  71.62 (49.62)\\ 
NUTS  &   47.4  &  >0.7  &  (7737.2, 20000.0, 20000.0)  &  166.93 (30.57)\\ 

  \bottomrule
  \end{tabular}
\end{table}

\section{Results on a higher dimensional example} 

We all tried a much larger Bayesian binary classification problem  by taking
all  $11339$ training examples of "5" and "6" MNIST digits which are $28 \times 28$ images
and therefore the dimensionality of the parameter vector $w$ was $785$ (the plus one accounts 
for the bias term). For this larger example from the baselines we applied the gradient-based 
schemes, MALA, HMC and NUTS since the other methods become very inefficient. 
From the proposed schemes we applied the most efficient algorithm which is gadMALAf. 
Also because of the much higher dimensionality of this problem, which makes the stochastic optimisation
over $L$ harder, we had to decrease the baseline learning rate in the RMSprop schedule from
$0.00015$ to $0.00001$. We also considered a larger adaptation phase consisted of $5 \times 10^4$
instead of $2 \times 10^4$. All other algorithms use the same experimental 
settings as described in the main paper.  Figure  \ref{fig:mnist} shows the evolution of the log-target
densities for all sampling schemes while Table  \ref{table:mnist} gives ESS, computation times and other statistics.  

We can observe that the performance of gadMALAf is reasonably good: it outperforms all methods apart form NUTS. 
 NUTS is better in this example, but it takes around 22 hours to run (since it performs 
on average 550 gradient evaluations per sampling iteration). Finally, to visualise some part of 
the learned $L$ found  by gadMALAf, Figure  \ref{fig:mnist} depicts the $784$ diagonal elements of $L$
as an $28 \times 28$ grey-scale image. Clearly, gadMALAf manages to perform a sort of feature selection, i.e.\ 
to discover that the border pixels in MNIST digits do not really take part in the classification, so it learns 
a much higher variance for those dimensions (close to the variance of the prior).   

\begin{figure}
\centering
\begin{tabular}{ccc}
\includegraphics[width=36mm,height=32mm]
{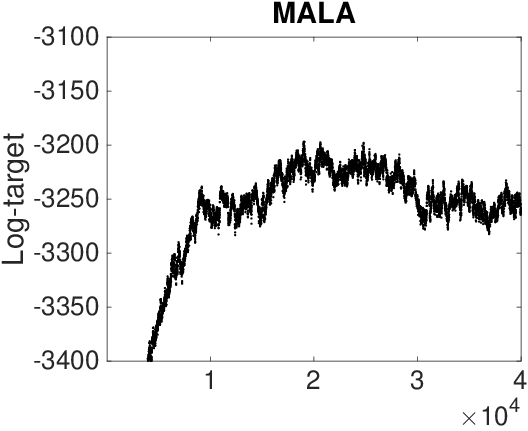} &
\includegraphics[width=36mm,height=32mm]
{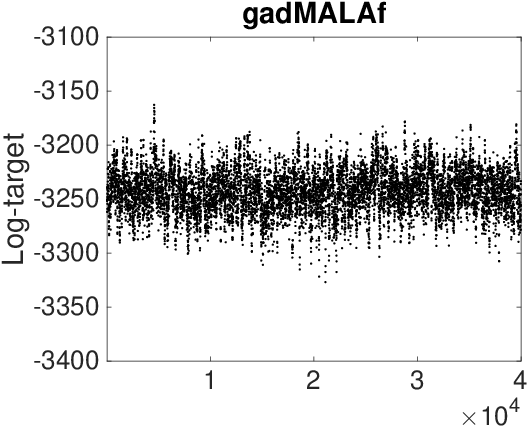}  &
\includegraphics[width=36mm,height=32mm]
{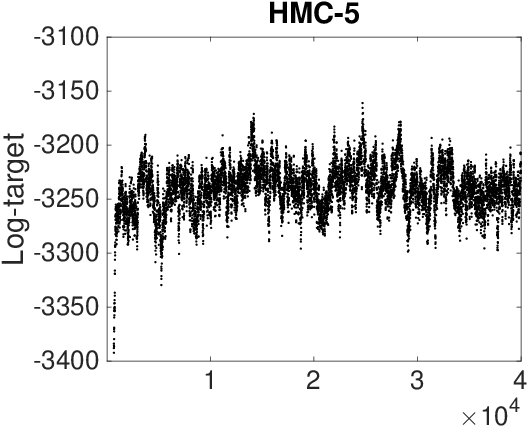} \\
\includegraphics[width=36mm,height=32mm]
{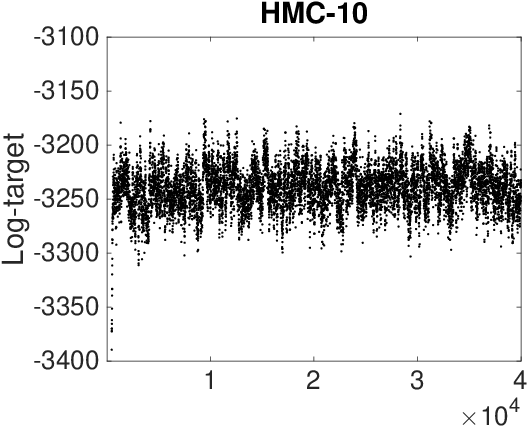} &
\includegraphics[width=36mm,height=32mm]
{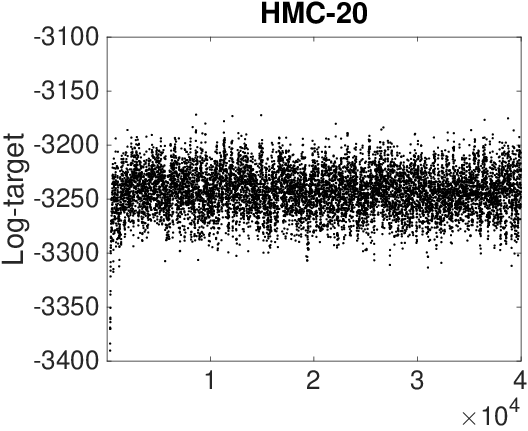} &
\includegraphics[width=36mm,height=32mm]
{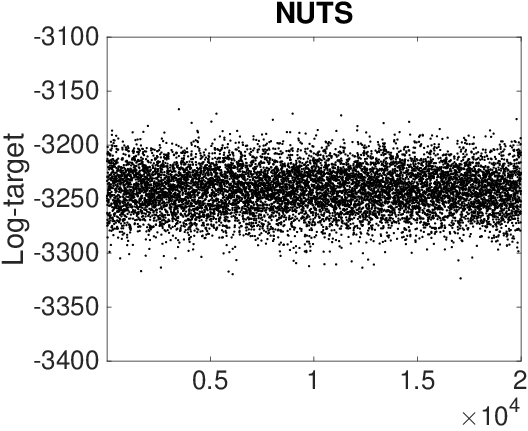} 
\end{tabular}
\caption{The evolution of  the log-target across iterations for all algorithms in binary MNIST classification 
over "5" versus "6".} 
\label{fig:mnist}
\end{figure}

\begin{table}
  \caption{Comparison of sampling methods in binary MNIST dataset, of "5" versus "6",  consisted of 
   $11339$   data points. The size of the state/parameter vector from which we draw samples was  $n = 785$.
All numbers are averages across five repeats where also one-standard deviation is given
for the Min ESS/s score.
  }
  \label{table:mnist}
  \centering
  \begin{tabular}{lllll}
    \toprule
 Method &  Time(s) & Accept Rate &  ESS (Min, Med, Max)  & Min ESS/s (1 st.d.) \\ 
\midrule
gadMALAf  &   779.3  &  0.575  &  (46.0, 128.7, 282.8)  &  0.059 (0.00)\\ 
MALA  &   311.8  &  0.530  &  (2.8, 5.9, 28.4)  &  0.009 (0.00)\\ 
HMC-5  &   1847.4  &  0.733  &  (4.5, 23.1, 162.7)  &  0.002 (0.00)\\ 
HMC-10  &   3381.3  &  0.589  &  (13.9, 66.5, 576.0)  &  0.004 (0.00)\\ 
HMC-20  &   6449.1  &  0.666  &  (77.8, 240.1, 2060.9)  &  0.012 (0.00)\\ 
NUTS  &   83232.1  &  >0.7  &  (18514.1, 20000.0, 20000.0)  &  0.223 (0.01)\\ 

  \bottomrule
  \end{tabular}
\end{table}

\begin{figure}
\centering
\begin{tabular}{c}
\includegraphics[width=90mm,height=90mm]
{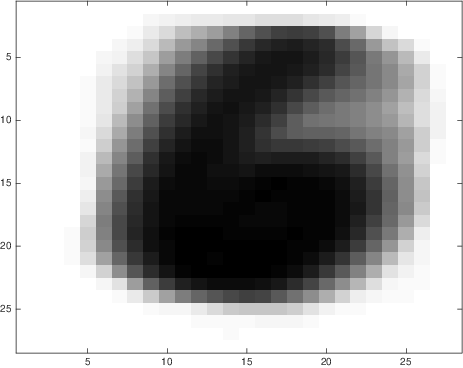} 
\end{tabular}
\caption{The first $784$ diagonal elements (i.e.\ excluding the bias component of $x$) 
of the full $785 \times 785$ Cholesky factor $L$ found after $5 \times 10^4$ adapting iterations by 
gadMALAf. Brighter/white colour  means larger values. } 
\label{fig:mnist}
\end{figure}


\end{document}